\documentclass[sigconf]{acmart}
\AtBeginDocument{%
  }

\setcopyright{acmlicensed}
\copyrightyear{2026}
\acmYear{2026}
\acmDOI{XXXXXXX.XXXXXXX}
\acmConference[Conference ACM MM'26]{Proceedings of the 34rd ACM International Conference on Multimedia}{November 10-14, 2026}{Rio de Janeiro, Brazil}
\acmBooktitle{Proceedings of the 34rd ACM International Conference on Multimedia (MM '26), November 10-14, 2026, Rio de Janeiro, Brazil}
\acmISBN{978-1-4503-XXXX-X/2018/06}

\usepackage{multirow}

\begin{document}

\title{UNIT: Unleash Large Language Models Potential for Graph Continual Learning}

\author{Tairan Huang}
\email{tairanhuang@csu.edu.cn}
\affiliation{%
  \institution{Central South University}
  \city{Changsha}
  \country{China}}

\author{Yili Wang}
\email{yili.wang@connect.hkust-gz.edu.cn}
\affiliation{%
  \institution{Hongkong University of Science and Technology (Guangzhou)}
  \city{Guangzhou}
  \country{China}}

\author{Beibei Hu}
\email{mindyhbb@gmail.com}
\affiliation{%
  \institution{Hunan Institute of Engineering}
  \city{Changsha}
  \country{Hunan}}

\author{Yiting Shi}
\email{shiyiting@csu.edu.cn}
\affiliation{%
  \institution{Central South University}
  \city{Changsha}
  \country{China}}

\author{Qiutong Li}
\email{qiutonglee@csu.edu.cn}
\affiliation{%
  \institution{Central South University}
  \city{Changsha}
  \country{China}}
  
\author{Changlong He}
\email{frankemail@csu.edu.cn}
\affiliation{%
  \institution{Central South University}
  \city{Changsha}
  \country{China}}

\author{Jianliang Gao}
\authornote{Corresponding author.}
\email{gaojianliang@csu.edu.cn}
\affiliation{%
  \institution{Central South University}
  \city{Changsha}
  \country{China}}
\renewcommand{\shortauthors}{Tairan Huang, Yili Wang, Beibei Hu, Yiting Shi, Qiutong Li, Changlong He, and Jianliang Gao}

\begin{abstract}
In real-world multimodal web scenarios, graph-structured data often arrives in a streaming manner, making graph continual learning a crucial paradigm for continuously modeling such evolving structures.
However, existing graph continual learning methods still face two fundamental challenges.
1) semantic-structural separation, where the graph-based methods excel at modeling topological relationships but neglect deep semantics.
2) imbalanced knowledge transfer, where existing models fail to effectively leverage general knowledge gained from early tasks to benefit subsequent new tasks.
To address above issues, we propose a novel framework, \textbf{UN}leash Large Language Models Potent\textbf{I}al for Graph Con\textbf{T}inual Learning (UNIT). 
By fine-tuning large language model only on the first task, we bridge the distributional gap between the pre-trained LLM corpus and the target task dataset to enhance the adaptability of LLMs for graph-structured tasks.
Meanwhile, we propose an uncertain-aware anchor generation mechanism to effectively preserve representative knowledge across tasks, avoiding the neglect of universal knowledge learned from previous tasks.
Additionally, we introduce structural confluence modeling to explicitly integrates graph topology information into semantic information, enhancing the collaborative capabilities between semantic understanding and structural modeling.
Extensive experiments demonstrate that our proposed method achieves state-of-the-art performance in the graph continual learning task.
\end{abstract}




\maketitle

\section{Introduction}
\begin{figure}[th]
	\centering
    \includegraphics[width=\linewidth]{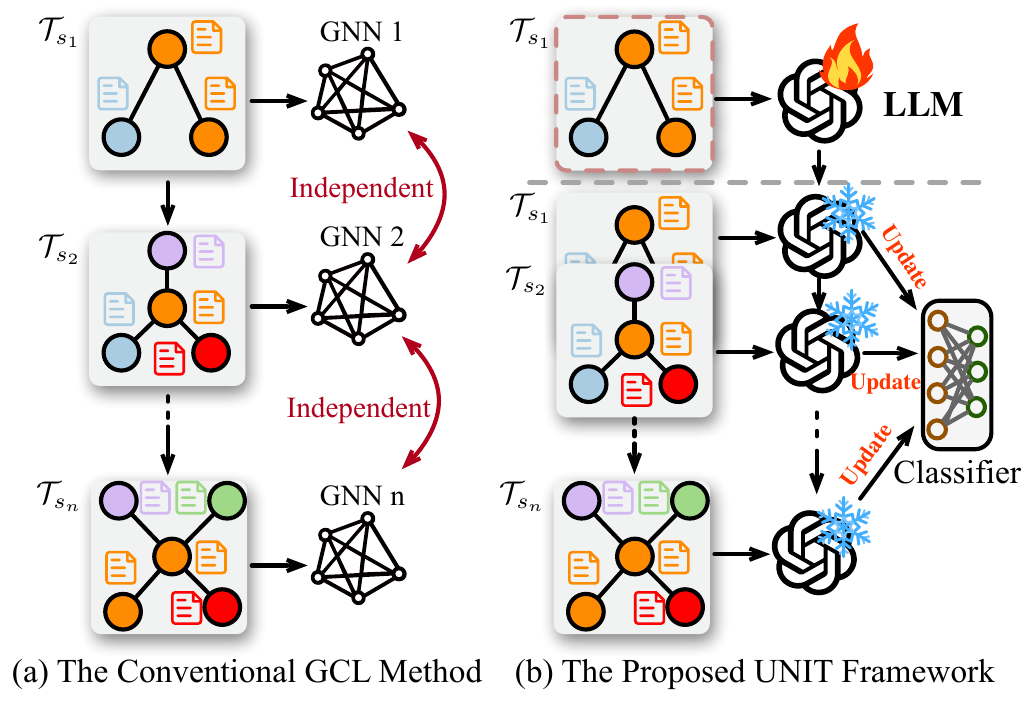}
	\caption{Workflow comparison. The existing graph continual learning methods need to train GNN models from scratch for each task.
    In contrast, our proposed framework only fine-tunes the LLM on the first task, while subsequent tasks update the classifier based on the frozen-parameter LLM.}
	\label{motivation}
\end{figure}

Graph Continual Learning (GCL) has become a crucial paradigm for modeling evolving multimodal data to analyze continuously changing structures in domains such as multimedia social networks \cite{social,social_2}, multimodal knowledge graphs \cite{Financial}, and multimedia recommendation systems \cite{Recommend}. 
With the strong capability to capture complex graph topologies alongside rich multimodal node attributes—such as text, images, and video—Graph Neural Networks (GNNs) have emerged as the dominant backbone for GCL tasks \cite{GCL}. 
Although existing efforts can preserve a portion of historical knowledge by regularization or empirical replay, they typically require retraining from scratch for each new task involving dynamic multimedia content \cite{SimGCL}, as shown in Figure \ref{motivation} (a). 
As a result, the model remains prone to overwriting previously learned cross-modal alignments and structural knowledge when adapting to new data streams, leading to catastrophic forgetting. 
Therefore, the emphasis on effective strategies to prevent catastrophic forgetting in dynamic multimedia environments remains the crucial challenge for graph continual learning tasks.

The existing graph continual learning approaches are broadly categorized into three paradigms based on their technical foundations.
The regularization-based methods \cite{Re,Re_2} achieve parameter regularization through explicit constraints via topological weighting or knowledge distillation.
The task-critical parameter isolation approaches \cite{is,is_2} achieve parameter isolation by extending parameters to accommodate new knowledge.
The replay-based approaches \cite{rh,rh_2} by replaying representative historical data, including key nodes, sparse computational subgraphs, and streamlined task-specific graphs, etc.
However, although existing graph continual learning methods alleviate catastrophic forgetting to some extent, they remain limited in effectively leveraging historical knowledge.
Specifically, replay-based methods face practical constraints, as privacy concerns and storage limitations often restrict access to historical task data, thereby undermining their applicability.
Therefore, several studies attempt to design different paradigms to mitigate catastrophic forgetting in graph continual learning tasks.

Recently, with the rapid advancement of large language models (LLMs), research interest in leveraging LLMs to model graph structures in conjunction with textual information has significantly increased, called Graph Language Models (GLMs) \cite{Roberta,SimpleCIL}.
The GLM methods typically leverage LLMs to retrieve node-related semantic knowledge for enhancing the quality of initial embeddings in GNNs \cite{GCN_emb,ENGINE}, or incorporate graph structural information into LLMs to improve classification predictions \cite{Graphgpt,Llaga}.
Although such methods can effectively integrate semantic and structural information, they still fail to fully exploit historical knowledge in graph continual learning.
In contrast, SimGCL \cite{SimGCL} is proposed as the first effort among GLM methods for graph continual learning tasks. 
It employs structured prompts to perform instruction-based fine-tuning on LLMs during the initial task, while leveraging LLMs for classification predictions in subsequent tasks.
Compared to general GLM methods, the graph continual learning tasks with GLMs remains largely unexplored.


Despite its directness and effectiveness, these methods mentioned above for graph continual learning task still exhibit several issues. 
First, existing methods suffer from \textit{structural information degradation} caused by the implicit linearization of graph topology. 
While SimGCL attempts to incorporate structural context via text prompts, they force non-Euclidean graph structures into linear sequences.
This linearization inevitably leads to the loss of high-order topological patterns and forces the LLM to infer structure solely through semantic attention, lacking an explicit geometric prior. 
Consequently, the distinct structural modality is overshadowed by dominant textual features, resulting in semantic-structural imbalance where critical topological changes in the evolving graph are underrepresented.
Second, current knowledge transfer mechanisms are prone to \textit{prototype rigidity}. 
SimGCL adopts a static prototype strategy that constructs class representations via simple mean pooling of sample embeddings. 
This approach assumes all samples are equally reliable, ignoring the inherent uncertainty and noise in streaming data.
In a continual setting, the deterministic aggregation fails to adaptively calibrate historical knowledge with new evidence, causing the model to overfit to noisy outliers in current tasks. 
This leads to suboptimal knowledge preservation and limits the model's plasticity for future tasks.

In this work, we propose a novel framework, \textbf{UN}leash Large Language Models Potent\textbf{I}al for Graph Con\textbf{T}inual Learning, i.e. UNIT. 
UNIT consists of three crucial components that \textit{effectively integrate semantic and structural information with efficient knowledge transfer}.
Specifically, the semantic instruction tuning module designs specific prompt instructions to fine-tune pre-trained LLMs, bridging the distributional gap between the pre-training corpus and the target task dataset to enhance the adaptability of LLMs to graph-structured tasks.
The uncertain-aware anchor generation utilizes fine-tuned LLMs to construct semantic anchors at each task, which updates classifier weights while achieves efficient knowledge transfer without forgetting previous knowledge.
Additionally, the structural confluence modeling explicitly generates structured anchors through graph topology, which jointly guide classifier weights with semantic anchors to achieve sufficient integration of semantic and structural information, enhancing the collaborative capabilities between semantic understanding and structural modeling.
The comprehensive experiments across five benchmark datasets demonstrate that UNIT can accurately classify the changing data from graph continual learning tasks to achieve state-of-the-art performance.
\begin{figure*}[th]
	\centering
	\includegraphics[width=\linewidth]{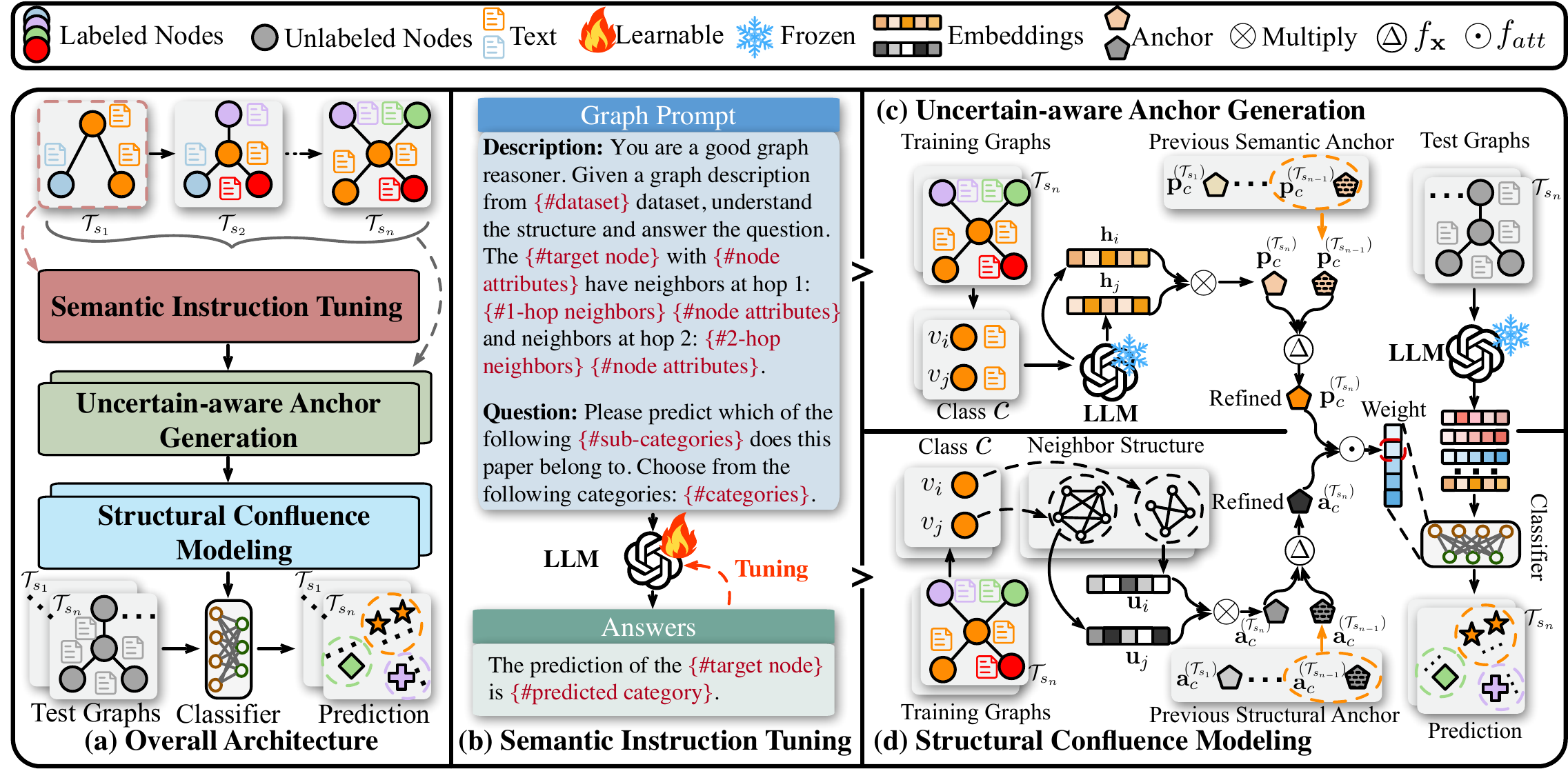}
	\caption{The overall framework of Unleash Large Language Models Potential for Graph Continual Learning (UNIT).
    The proposed framework includes three stages, i.e., the semantic instruction tuning, the uncertain-aware anchor generation, and the structural confluence modeling.
    The pre-trained LLM is fine-tuned by specific instruction prompts, then generates semantic anchors and structural anchors to jointly update classifier weights, achieving accurate classification prediction.
    }
	\label{framework}
\end{figure*}

Our contributions are summarized as follows:
\begin{itemize}
    
	\item \textbf{Specific Design.} 
    We deign specific instruction prompts at the initial task on LLMs for fine-tuning to bridge the distributional gap between the pre-trained LLM corpus and the target task dataset, facilitating LLM adaptation to graph continual learning tasks.
    
	\item \textbf{Novel framework.} 
    We propose the uncertain-aware anchor generation and the structural confluence modeling to achieve comprehensive knowledge transfer and the sufficient integration of semantic information and structural information for graph continual learning tasks.
    
	\item \textbf{SOTA performance.}  
    We conduct extensive experiments on five real-world datasets to demonstrate the effectiveness of UNIT, achieving state-of-the-art performance on graph continual learning tasks.
\end{itemize}

\section{Related Works}
\subsection{Continual Learning with LLMs}
Continual learning has long been studied as a fundamental paradigm for training models that incrementally adapt to new tasks without forgetting previously acquired knowledge \cite{GCL}, which as become a crucial paradigm for modeling evolving multimodal data.
With the rapid rise of large language models, an increasing number of studies are exploring how to extend these models to continual learning scenarios \cite{BERT,llama,lora,Llaga}.
The language model for continual learning task \cite{few_1} exhibits inherent advantages in generalization capabilities when deployed for continuous learning tasks after being pre-trained on extensive corpora.
InfLoRA \cite{Inflora} and BiLoRA \cite{BiLoRA} employ parameter-efficient fine-tuning (PEFT) through methods such as freezing the backbone or retaining pre-trained weights.
ENGINE \cite{ENGINE} and SimpleCIL \cite{SimpleCIL} minimize cross-task interference by adopting orthogonal optimizations.
However, despite these advances, continual learning with LLMs focus on preventing catastrophic forgetting of semantic representations, remaining under-explored for graph continual learning tasks.
Instead, our method only fine-tunes the LLM with the specific prompt in the initial task and completes subsequent tasks using the LLM with frozen parameters.

\subsection{Graph Continual Learning}
Graph continual learning primarily investigates catastrophic forgetting arising from evolving graph structures \cite{GCL}.
The graph continual learning methods can be categorized based on their approaches to mitigating catastrophic forgetting.
CPG \cite{Re} and ER-GNN \cite{Re_2} serve as regularization-based methods, which impose explicit constraints on the model through parameter regularization via topology-based weighting or knowledge distillation.
PI-GNN \cite{is} and HPNs\cite{is_2} are proposed as key parameter isolation methods, achieving parameter isolation through expanding parameters or model components to accommodate new knowledge.
SSM \cite{rh}, SEM\cite{rh_2}, and TPP \cite{TPP} act as replay-based methods, which replay representative historical data using key nodes and sparse computational subgraphs respectively to prevent catastrophic forgetting.
However, these methods face significant limitations in leveraging historical knowledge due to scalability and privacy concerns in the field of graphs.
In contrast, we introduce uncertain-aware anchor generation for efficient knowledge transfer, which can leverage historical knowledge through LLMs to inspirit future tasks.

\subsection{Graph Language Models}
Graph language models have gradually emerged as popular approaches for leveraging LLMs to address graph tasks \cite{BERT,Roberta,llama,Llaga}, driven by the rapid advancement of LLMs.
Based on how LLMs are integrated with graph structures, GLMs methods can be categorized into three categories.
Glbench \cite{Glbench} and Graphtranslator \cite{Graphtranslator} utilize LLMs as enhancers to generate semantic representations for node textual information, which improves the quality of GNN node embeddings.
TEA-GLM \cite{TEA-GLM} and MAGRL \cite{MAGRL} act LLMs as predictor to perform classification predictions after designing specific prompts for graph structural information.
CoT \cite{CoT} and Locle \cite{Locle} serve as alignment methods, which allow LLMs and GNNs to be aligned within the same vector space, providing GNNs with clearer semantic awareness.
These GLM methods are not specifically designed for continual graph data, whose application in graph continual learning remains largely unexplored.
Beyond these models, SimGCL \cite{SimGCL} is proposed as the first GLM method for graph-continual learning tasks. 
It employs a static prototype strategy to guide classification predictions for the latest data, exhibiting forgetting of early information and suboptimal adaptation to future tasks.
Compared to SimGCL, our work achieves efficient knowledge transfer and fully integrates semantic and structural information through uncertain-aware anchor generation and structural confluence modeling.

\section{Methodology}
\subsection{Preliminaries}
\noindent{\bfseries Text-attributed Graphs (TAGs).}
According to LLM4GCL benchmark \cite{SimGCL}, we evaluate the graph continual learning capabilities through node classification tasks on Text-attributed Graphs (TAGs) \cite{TAG}.
The text-attributed graph is defined as $\mathcal{G = \{V, E, R\}}$, where $\mathcal{V} = \{v_1, v_2, \ldots, v_N\}$ is the set of nodes, $\mathcal{E}$ indicates the edge set, and $\mathcal{R}$ denotes the collection of textual descriptions. 
Each node $v_i \in \mathcal{V}$ is associated with a text description $r_{v_i} \in \mathcal{R}$, drawn from the vocabulary $\mathcal{D}$ and of length $L_i$. 
Given a TAG with a set of labeled nodes $\mathcal{V}^l$, the node classification objective is to infer the labels of the unlabeled nodes $\mathcal{V}^u$.

\noindent{\bfseries Graph Continual Learning.}
The fundamental paradigm in graph continual learning is defined as node-level class-incremental learning \cite{SimGCL}, that tackles the challenge of progressively classifying nodes as new categories appear in an expanding graph $\mathcal{G}$.
Formally, the training procedure is organized as a sequence of tasks $\{\mathcal{T}_{s_1}, \ldots, \mathcal{T}_{s_n}\}$, with each task $\mathcal{T}_{s_i} = \{\mathcal{G}_{s_i}, \mathcal{C}_{s_i}, \mathcal{V}^l_{s_i}\}$, where $\mathcal{G}_{s_i} \subseteq \mathcal{G}$ is the induced subgraph, $\mathcal{C}_{s_i}$ the corresponding class set, and $\mathcal{V}^l_{s_i}$ the labeled nodes. 
The disjointness constraint holds such that $\mathcal{G}_{s_i} \cap \mathcal{G}_{s_j} = \mathcal{C}_{s_i} \cap \mathcal{C}_{s_j} = \varnothing$ for all $i \neq j$. 
During learning, the model incrementally adapts to each arriving task $\mathcal{T}_{s_i}$ and predicts the labels of the unseen nodes.

\subsection{Semantic Instruction Tuning}
It is crucial to design specific instruction prompts for tuning pre-trained LLMs to unleash their potential for graph continual learning.
The semantic instruction tuning facilitates LLM adaptation to graph-structured tasks by bridging the distributional gap between the pre-trained LLM corpus and the target task dataset.
Specifically, we construct the comprehensive graph prompt that encapsulates: (1) the task description with the continual learning scenario and the target dataset; 
(2) the structural information of the graph, including node attributes and local neighborhood connectivity;
(3) the task-specific question that defines the classification objective along with the set of candidate label descriptions.
Formally, for a node $v_i \in \mathcal{V}$ with textual attribute $r_{v_i}$ and neighbor set $\mathcal{N}_{v_i}$, the prompt is structured as:
\begin{equation}
    \text{Prompt}(v_i) = r_{\text{task}} \oplus r_{\text{str}}(v_i) \oplus r_{\text{que}}(v_i),
\end{equation}
where $r_{\text{task}}$ describes the task description, $r_{\text{str}}(v_i)$ summarizes the node attributes and local structures of node $v_i$, and $r_{\text{que}}(v_i)$ specifies the classification query along with the label text options. 
This prompt design enables the LLM to leverage semantic and structural cues for classification.
To prevent overfitting and maintain efficiency, we perform instruction tuning only on the first task $\mathcal{T}_{s_1}$ using Low-Rank Adaptation (LoRA) \cite{lora} rather than full fine-tuning. 
Formally, given a node $v_i$ with prompt $\text{Prompt}(v_i)$, the objective is to maximize the correct answer probability with label token sequence $q_i = (q_{i,1}, \ldots, q_{i,L})$. 
The answer probability defined by the LLM with parameters $\theta$ can be expressed as:
\begin{equation}
\max_{\theta} \; \mathcal{P}_{\theta}\!\left(q_i \mid v_i\right) 
= \max_{\theta} \; \prod_{l=1}^{L} \mathcal{P}_{\theta}\!\left(q_{i,l} \mid q_{i,<l}, \text{Prompt}(v_i)\right),
\end{equation}
where $q_{i,l}$ denotes the $l$-th token of the target label text, and $q_{i,<l}$ represents its preceding tokens. 
Here, $\text{Prompt}(v_i)$ encapsulates the task description, the node textual attribute, the neighbor context, and the query question. 
The probability $\mathcal{P}_{\theta}$ is modeled by the instruction-tuned LLM with LoRA parameters $\theta$, which ensures efficient adaptation without full-scale fine-tuning. 
This formulation explicitly decomposes the sequence-level likelihood into token-level conditionals, encouraging the model to align its generative distribution with the semantic and structural cues provided in the constructed prompt.
In Figure \ref{framework}, we provide an overview of UNIT.

\subsection{Uncertain-aware Anchor Generation}
To effectively preserve representative knowledge across tasks, we design an uncertain-aware anchor generation mechanism that progressively constructs class-level semantic anchors while retaining the influence of early samples. 
We generate the semantic anchors for labeled nodes in each task to update the weights of corresponding classes in the classifier, thereby improving the accuracy of classification prediction.

Specifically, for each labeled node $v_i \in \mathcal{V}^l_{s_i}$ in task $\mathcal{T}_{s_i}$, we obtain its embedding representation through the tuned LLM encoder. 
Given the constructed prompt $\text{Prompt}(v_i)$, the node embedding extraction is defined as:
\begin{equation}
\mathbf{h}_i = f_{\theta}\!\left(\text{Prompt}(v_i)\right),
\end{equation}
where $f_{\theta}$ denotes the embedding function parameterized by the LoRA-tuned LLM. 
These embeddings serve as the basis for anchor construction.  

Then, to compute a reliable anchor for each class $c \in \mathcal{C}_{s_i}$, we weight individual embeddings according to their uncertainty-aware confidence.
Let $\hat{p}(c|v_i)$ be the normalized probability that the LLM predicts class $c$ for node $v_i$, derived from the generation scores. 
The uncertainty is quantified through the entropy:
\begin{equation}
\mathcal{H}(v_i) = - \sum_{c} \hat{p}(c|v_i) \log \hat{p}(c|v_i).
\end{equation}
Consequently, the confidence weight $\alpha_i \in [0,1]$ is explicitly defined as $\alpha_i = 1 - \frac{\mathcal{H}(v_i)}{\log |\mathcal{C}_{s_i}|}$, where $|\mathcal{C}_{s_i}|$ is the number of classes. 
The class-level anchor is computed as:
\begin{equation}
\mathbf{p}_c^{(\mathcal{T}_{s_i})} = \frac{\sum_{y_i \in \mathcal{Y}_c^{(\mathcal{T}_{s_i})}} \alpha_i \mathbf{h}_i}{\sum_{y_i \in \mathcal{Y}_c^{(\mathcal{T}_{s_i})}} \alpha_i},
\end{equation}
where $\mathcal{Y}_c^{(\mathcal{T}_{s_i})}$ is the label set of nodes in task $\mathcal{T}_{s_i}$ belonging to class $c$. 
In parallel, we maintain an effective count $n_c^{(\mathcal{T}_{s_i})}$ that records the accumulated contribution of class-$c$ samples up to task $\mathcal{T}_{s_i}$, which is updated as:
\begin{equation}
n_c^{(\mathcal{T}_{s_i})} = \lambda \cdot n_c^{(\mathcal{T}_{s_{i-1}})} + \sum_{y_i \in \mathcal{Y}_c^{(\mathcal{T}_{s_i})}} \alpha_i,
\end{equation}
where $\lambda \in (0,1]$ is the exponential decay factor. 
This weighted count ensures that reliable samples have higher impact while outdated contributions gradually fade.  

Next, we update the class anchor by softly merging the newly aggregated embedding with the decayed historical anchor. 
Let $\mathbf{p}_c^{(\mathcal{T}_{s_{i-1}})}$ be the anchor obtained after task $\mathcal{T}_{s_{i-1}}$. The refined anchor at task $\mathcal{T}_{s_i}$ is given by:
\begin{equation}
\mathbf{p}_c^{(\mathcal{T}_{s_i})} = \frac{\lambda \cdot n_c^{(\mathcal{T}_{s_{i-1}})} \mathbf{p}_c^{(\mathcal{T}_{s_{i-1}})} + \sum_{y_i \in \mathcal{Y}_c^{(\mathcal{T}_{s_i})}} \alpha_i \mathbf{h}_i}{n_c^{(\mathcal{T}_{s_i})}},
\end{equation}
which softly discounts early anchors while adaptively emphasizing reliable new evidence.  

Finally, the refined semantic anchors are combined with subsequent structural anchors and aligned with classifier weights, primarily guiding the decision boundary toward the most representative class semantics.
By embedding anchors into the classifier, we seamlessly bridge semantic representations and task-specific predictions, enabling the model to achieve effective knowledge transfer across tasks.
Overall, the uncertain-aware anchor generation mechanism progressively refines semantic anchors to learn knowledge from new samples while retaining the influence of early samples, enhancing the capabilities graph continual learning.

\begin{table*}
\setlength{\tabcolsep}{3.7pt}
\centering
\caption{Performance comparison of UNIT and baselines. 
}
\begin{tabular}{ccccccccccc}
\toprule
\multirow{2}{*}{Methods} & \multicolumn{2}{c}{Cora} & \multicolumn{2}{c}{Citeseer} & \multicolumn{2}{c}{WikiCS} & \multicolumn{2}{c}{Photo} & \multicolumn{2}{c}{Products} \\
\cmidrule(lr){2-3} \cmidrule(lr){4-5} \cmidrule(lr){6-7} \cmidrule(lr){8-9} 
\cmidrule(lr){10-11} 
 & ACC$_{\text{avg}}$         & ACC$_\text{N}$        & ACC$_{\text{avg}}$         & ACC$_\text{N}$          & ACC$_{\text{avg}}$         & ACC$_\text{N}$        & ACC$_{\text{avg}}$         & ACC$_\text{N}$           & ACC$_{\text{avg}}$         & ACC$_\text{N}$     \\
\midrule
GCN    & 57.0 ± 0.6 & 38.2 ± 1.2 & 52.4 ± 1.9 & 30.2 ± 0.6 & 54.9 ± 1.8 & 34.9 ± 0.9 & 46.5 ± 1.3 & 19.9 ± 1.7 & 25.5 ± 0.8 & 5.4 ± 1.5 \\
EWC    & 56.0 ± 1.9 & 31.0 ± 0.7 & 45.9 ± 1.6 & 28.5 ± 1.8 & 55.3 ± 0.6 & 33.0 ± 1.3 & 46.9 ± 0.6 & 20.5 ± 2.0 & 29.4 ± 1.1 & 15.9 ± 0.7 \\
LwF    & 55.7 ± 1.4 & 30.8 ± 0.8 & 47.8 ± 1.6 & 28.2 ± 1.3 & 55.0 ± 2.0 & 34.0 ± 0.7 & 45.8 ± 1.1 & 20.4 ± 1.5 & 26.0 ± 0.6 & 7.5 ± 1.8 \\
Cosine & 65.4 ± 0.7 & 45.2 ± 1.4 & 50.7 ± 1.7 & 31.0 ± 0.8 & 66.5 ± 1.2 & 53.5 ± 0.9 & 63.6 ± 1.8 & 49.6 ± 1.0 & 36.1 ± 1.6 & 16.1 ± 1.4 \\
TPP    & 45.7 ± 1.2 & 13.7 ± 0.9 & 45.4 ± 0.5 & 9.6 ± 1.7 & 35.1 ± 1.5 & 9.8 ± 1.0 & 36.3 ± 0.8 & 5.7 ± 1.9 & 15.0 ± 1.2 & 0.2 ± 0.3 \\
\midrule
BERT      & 56.0 ± 1.3 & 29.9 ± 0.9 & 53.8 ± 1.4 & 28.7 ± 1.6 & 58.4 ± 0.7 & 30.0 ± 1.2 & 43.4 ± 1.9 & 18.4 ± 0.8 & 26.9 ± 1.3 & 4.1 ± 1.6 \\
RoBERTa   & 54.6 ± 1.7 & 29.6 ± 1.1 & 54.1 ± 0.6 & 28.6 ± 1.8 & 55.1 ± 1.8 & 30.8 ± 0.7 & 43.5 ± 1.1 & 19.1 ± 1.4 & 27.6 ± 1.0 & 3.2 ± 1.3 \\
LLaMA     & 65.6 ± 0.8 & 53.8 ± 1.6 & 55.7 ± 2.0 & 31.7 ± 0.9 & 55.5 ± 1.4 & 30.9 ± 1.1 & 44.6 ± 0.7 & 19.1 ± 1.7 & 29.8 ± 1.5 & 0.4 ± 0.2 \\
SimpleCIL & 70.8 ± 1.9 & 58.3 ± 1.0 & 66.4 ± 1.3 & 49.5 ± 0.7 & 71.4 ± 1.5 & 57.3 ± 1.2 & 62.1 ± 0.8 & 52.5 ± 1.5 & 66.8 ± 1.8 & 52.6 ± 0.7 \\
\midrule
GCN$_{\text{LLMEmb}}$   & 59.1 ± 1.5 & 31.1 ± 1.7 & 53.6 ± 1.0 & 30.4 ± 1.1 & 53.4 ± 1.6 & 27.5 ± 0.9 & 47.7 ± 1.0 & 21.0 ± 1.5 & 26.9 ± 1.9 & 0.1 ± 0.2 \\
ENGINE      & 59.2 ± 1.2 & 31.3 ± 1.4 & 53.5 ± 1.8 & 29.8 ± 0.7 & 56.4 ± 0.9 & 30.1 ± 1.9 & 47.9 ± 1.7 & 21.0 ± 1.1 & 27.2 ± 0.8 & 1.1 ± 0.4 \\
GraphPrompter & 61.9 ± 1.0 & 46.8 ± 1.5 & 60.2 ± 1.3 & 30.6 ± 1.8 & 59.6 ± 1.7 & 38.3 ± 0.8 & 51.5 ± 1.4 & 31.0 ± 0.9 & 29.0 ± 1.6 & 0.8 ± 0.3 \\
GraphGPT    & 55.5 ± 1.7 & 31.6 ± 1.0 & 60.0 ± 1.4 & 30.1 ± 1.3 & 62.0 ± 0.9 & 49.2 ± 1.6 & 50.8 ± 1.2 & 30.2 ± 1.8 & 35.5 ± 2.0 & 3.2 ± 0.5 \\
LLaGA       & 58.2 ± 1.6 & 30.2 ± 1.9 & 51.3 ± 0.7 & 27.8 ± 1.4 & 53.7 ± 1.3 & 27.6 ± 1.7 & 47.2 ± 1.8 & 20.7 ± 0.9 & 25.7 ± 1.1 & 0.2 ± 0.1 \\
SimGCL  & \underline{84.6 ± 0.5} & \underline{80.0 ± 2.0} & \underline{77.1 ± 1.3} & \underline{66.3 ± 0.8} & \underline{73.5 ± 1.9} & \underline{61.9 ± 1.1} & \underline{82.1 ± 1.2} & \underline{72.6 ± 0.9} & \underline{71.1 ± 1.6} & \underline{60.2 ± 1.1} \\
\midrule
\textbf{UNIT (ours) } & \textbf{88.7 ± 0.4} & \textbf{85.9 ± 1.0} & \textbf{79.4 ± 0.7} & \textbf{68.5 ± 0.9} & \textbf{76.1 ± 0.3} & \textbf{65.0 ± 0.7} & \textbf{83.9 ± 1.1} & \textbf{76.5 ± 0.7} & \textbf{73.2 ± 1.1} & \textbf{62.5 ± 1.3} \\
\bottomrule
\end{tabular}
\label{main_results}
\end{table*}

\subsection{Structural Confluence Modeling}
To further enrich the representation capacity of anchors, we introduce structural confluence modeling, which explicitly integrates graph topology into the anchor space. 
While semantic instruction tuning and uncertain-aware anchor generation focus on textual and confidence-weighted representations, this module complements them by encoding lightweight structural cues that capture local connectivity patterns. 
This confluence ensures that both semantic and structural perspectives jointly contribute to stable classification boundaries under continual learning, resolving the imbalance
between semantic understanding and structural modeling.  

First, for the node $v_i \in \mathcal{V}^l_{s_i}$, we derive a structural embedding $\mathbf{u}_i$ that summarizes the topological characteristics of its neighborhood.
Let $\mathcal{N}(v_i)$ denote the neighbor set of $v_i$.
We represent structural information as follows:
\begin{equation}
\mathbf{u}_i = g\!\left(d_i, \; \frac{1}{|\mathcal{N}(v_i)|}\sum_{k \in \mathcal{N}(v_i)} d_k, \; \phi(\mathcal{N}(v_i)) \right),
\end{equation}
where $d_i = |\mathcal{N}(v_i)|$ is the degree of $v_i$, the second term encodes average neighbor degree, and $\phi(\mathcal{N}(v_i))$ is the clustering coefficient capturing local density. 
The function $g(\cdot)$ projects quantities into a fixed-dimensional embedding space.  

Second, for each class $c$ in the active class set $\mathcal{C}_{s_t}$ of task $\mathcal{T}_{s_i}$, we maintain a structural anchor by aggregating the descriptors of its labeled nodes. 
Formally, given the labeled set $\mathcal{Y}_c^{(\mathcal{T}_{s_i})} = \{ v_i \mid y_i = c \}$, the temporary anchor is defined as:
\begin{equation}
\mathbf{a}_c^{(\mathcal{T}_{s_i})} = \frac{1}{|\mathcal{Y}_c^{(\mathcal{T}_{s_i})}|} \sum_{y_i \in \mathcal{Y}_c^{(\mathcal{T}_{s_i})}} \mathbf{u}_i.
\end{equation}
This aggregation produces a representative vector that captures the structural tendencies of class $c$ in the current task.  

Third, to ensure stability across tasks and avoid abrupt shifts, we update the structural anchors with the exponential decay, paralleling the semantic anchor update. 
Let $\mathbf{a}_c^{(t-1)}$ be the structural anchor of class $c$ from the task $t-1$, and let $m_c^{(t-1)}$ be its effective count. After observing task $t$, the anchor is updated as:
\begin{equation}
\mathbf{a}_c^{(\mathcal{T}_{s_i})} = \frac{\gamma \cdot m_c^{(\mathcal{T}_{s_{i-1}})} \mathbf{a}_c^{(\mathcal{T}_{s_{i-1}})} + |\mathcal{Y}_c^{(\mathcal{T}_{s_i})}| \cdot \mathbf{a}_c^{(\mathcal{T}_{s_i})}}{\gamma \cdot m_c^{(\mathcal{T}_{s_{i-1}})} + |\mathcal{Y}_c^{(\mathcal{T}_{s_i})}|},
\end{equation}
with the effective count maintained as:
\begin{equation}
m_c^{(\mathcal{T}_{s_i})} = \gamma \cdot m_c^{(\mathcal{T}_{s_{i-1}})} + |\mathcal{Y}_c^{(\mathcal{T}_{s_i})}|,
\end{equation}
where $\gamma \in (0,1]$ is the decay factor. 
This formulation allows the model to gradually discount outdated structural evidence while adapting to new patterns.  

Finally, we injects the structural anchors into the classifier alongside the semantic anchors obtained from uncertain-aware anchor generation.
Specifically, the classifier weight for class $c$ is updated as a convex combination of the semantic and structural anchors:
\begin{equation}
\mathbf{w}_c^{(\mathcal{T}_{s_i})} = \beta \cdot \mathbf{p}_c^{(\mathcal{T}_{s_i})} + (1-\beta)\cdot \mathbf{a}_c^{(\mathcal{T}_{s_i})},
\end{equation}
where $\mathbf{p}_c^{(\mathcal{T}_{s_i})}$ is the semantic anchor produced by the uncertain-aware anchor generation mechanism, $\mathbf{a}_c^{(\mathcal{T}_{s_i})}$ is the structural anchor, and $\beta \in [0,1]$ balances semantic and structural contributions.  
Overall, the structural confluence modeling module enriches the anchor space by combining semantic reasoning with structural confluence.
The combined use of textual and topological anchors enables the model to leverage complementary signals by explicitly integrating graph topology into semantic representations, mitigating the imbalance between semantic understanding and structural modeling.

For the classification inference, each test node $v_j \in \mathcal{V}^u$ is represented by its semantic embedding $\mathbf{h}_j=f_\theta(\text{Prompt}(v_j))$, and classification is performed through the updated classifier:
\begin{equation}
z_{j,c} = \langle \mathbf{h}_j, \mathbf{w}_c^{(\mathcal{T}_{s_i})} \rangle,
\qquad
\hat{y}_j = \arg\max_c z_{j,c}.
\end{equation}

\section{Experiments}

\subsection{Experimental Setup}

\begin{figure*}[th]
	\centering
    \includegraphics[width=\linewidth]{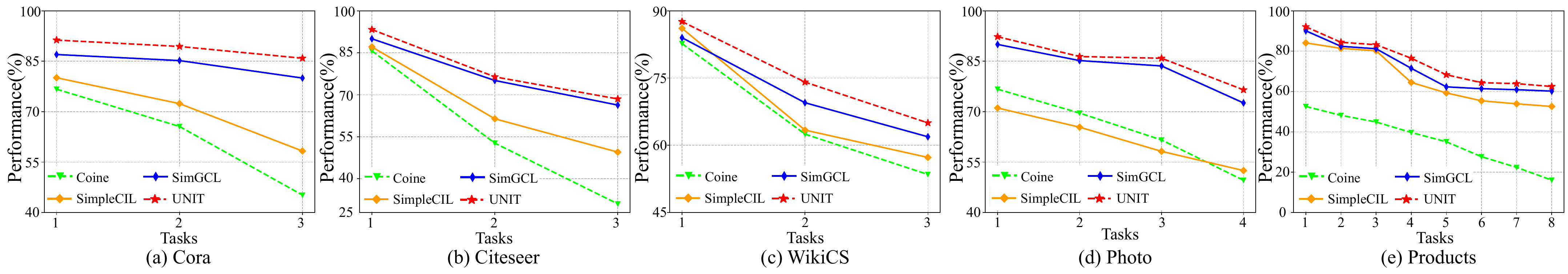}
	\caption{The performance comparison between UNIT and state-of-the-art methods in three baseline categories with ACC$_\text{N}$.
    }
	\label{task}
\end{figure*}

\noindent{\bfseries Datasets.}
We evaluate the proposed UNIT and all baseline methods on five benchmark datasets including Cora \cite{Cora}, Citeseer \cite{Citeseer}, WikiCS \cite{WikiCS}, Photo \cite{Photo}, and Products \cite{Products}.
We follow the settings of LLM4GCL benchmark \cite{SimGCL} by dividing the datasets into multiple tasks and classes per task to conduct a comprehensive analysis of the model's continual learning performance.

\noindent{\bfseries Baselines.}
We conduct a comprehensive comparison of UNIT with fifteen baseline methods, which fall into three categories.
The GNN-based methods include GCN \cite{GCN}, EWC \cite{EWC}, LwF \cite{LwF}, Cosine \cite{Cosine}, and TPP \cite{TPP}.
The LLM-based methods include encoder-only BERT \cite{BERT}, RoBERTa \cite{Roberta}, decoder-only LLaMA \cite{llama}, and SimpleCIL \cite{SimpleCIL}.
The GLM-based methods include two LLM-as-Enhancer methods GCN$_{\text{LLMEmb}}$ \cite{GCN_emb} and ENGINE \cite{ENGINE}, and three LLM-as-Predictor methods GraphPrompter \cite{GraphPrompter}, GraphGPT \cite{Graphgpt}, LLaGA \cite{Llaga}, and SimGCL \cite{SimGCL}.

\noindent{\bfseries Metrics.}
We employ accuracy as the evaluation metric for the node classification task in graph continual learning.
Following the standard setting established continual learning benchmark \cite{GCL_bench}, we define ACC$_{i}$ as the accuracy after the $i$-th training task.
We adopt the final metrics ACC$_{\text{avg}}=\frac{1}{N} {\textstyle \sum_{i=1}^{N}} \text{ACC}_{i}$ and ACC$_\text{N}$, which refer to the average accuracy across all tasks and the accuracy of the last task, respectively.

\noindent{\bfseries Implementations.}
To ensure fairest possible comparison across LLM-driven baseline methods, we use same LLM backbone \emph{LLaMA3-8B} in main experiment.
The hyperparameters decay factors $\lambda$ and $\gamma$ are set to the size of $\lambda=0.7$ and $\gamma=0.6$, respectively.

\begin{table*}
\setlength{\tabcolsep}{3.7pt}
\centering
\caption{Performance comparison of UNIT and baselines in the \textbf{\textit{few-shot}} scenarios. 
}
\begin{tabular}{ccccccccccc}
\toprule
\multirow{2}{*}{Methods} & \multicolumn{2}{c}{Cora} & \multicolumn{2}{c}{Citeseer} & \multicolumn{2}{c}{WikiCS} & \multicolumn{2}{c}{Photo} & \multicolumn{2}{c}{Products} \\
\cmidrule(lr){2-3} \cmidrule(lr){4-5} \cmidrule(lr){6-7} \cmidrule(lr){8-9} 
\cmidrule(lr){10-11} 
 & ACC$_{\text{avg}}$         & ACC$_\text{N}$        & ACC$_{\text{avg}}$         & ACC$_\text{N}$          & ACC$_{\text{avg}}$         & ACC$_\text{N}$        & ACC$_{\text{avg}}$         & ACC$_\text{N}$           & ACC$_{\text{avg}}$         & ACC$_\text{N}$    \\
\midrule
GCN    & 68.0 ± 0.5 & 38.1 ± 1.3 & 39.5 ± 1.2 & 17.4 ± 0.4 & 62.4 ± 1.2 & 47.4 ± 0.4 & 58.5 ± 1.6 & 32.3 ± 1.2 & 36.0 ± 0.9 & 14.1 ± 1.0 \\
EWC    & 59.0 ± 1.4 & 36.3 ± 0.8 & 49.2 ± 1.4 & 21.2 ± 1.3 & 58.4 ± 0.8 & 40.4 ± 1.2 & 62.0 ± 0.7 & 28.4 ± 1.7 & 45.7 ± 1.3 & 31.5 ± 0.5 \\
LwF    & 63.3 ± 1.6 & 43.5 ± 0.2 & 45.1 ± 1.3 & 20.7 ± 1.6 & 59.4 ± 1.6 & 41.0 ± 0.4 & 60.1 ± 1.2 & 29.4 ± 1.3 & 50.3 ± 0.7 & 38.7 ± 1.6 \\
Cosine & 72.6 ± 0.8 & 57.8 ± 1.3 & 49.1 ± 1.6 & 25.7 ± 0.7 & 68.0 ± 1.3 & 50.7 ± 0.7 & 67.9 ± 1.4 & 50.5 ± 1.2 & 50.9 ± 1.3 & 33.6 ± 1.5 \\
TEEN   & 60.9 ± 1.3 & 40.3 ± 0.7 & 59.0 ± 0.7 & 39.5 ± 1.8 & 59.3 ± 1.3 & 42.4 ± 1.2 & 59.3 ± 0.9 & 35.5 ± 1.7 & 49.6 ± 1.1 & 28.4 ± 1.8 \\
TPP    & 39.0 ± 1.1 & 9.1 ± 0.7 & 37.3 ± 1.2 & 12.6 ± 1.3 & 37.3 ± 0.6 & 12.4 ± 1.0 & 40.0 ± 1.2 & 14.0 ± 0.9 & 14.0 ± 1.2 & 0.2 ± 0.1 \\
\midrule
BERT   & 56.4 ± 1.5 & 34.7 ± 1.2 & 61.1 ± 0.7 & 38.8 ± 1.7 & 61.7 ± 1.9 & 33.0 ± 0.9 & 47.7 ± 1.2 & 25.8 ± 1.5 & 22.4 ± 1.1 & 3.8 ± 1.0 \\
RoBERTa & 59.6 ± 0.7 & 41.9 ± 1.5 & 54.0 ± 1.8 & 29.2 ± 0.8 & 67.2 ± 1.3 & 42.3 ± 1.2 & 58.2 ± 0.8 & 29.6 ± 1.5 & 38.8 ± 1.4 & 6.9 ± 1.1 \\
LLaMA  & 72.6 ± 1.7 & 55.6 ± 1.1 & 75.9 ± 1.2 & 55.5 ± 0.8 & 65.2 ± 1.3 & 43.9 ± 1.4 & 61.4 ± 0.9 & 32.3 ± 1.2 & 43.5 ± 1.6 & 13.7 ± 0.8 \\
SimpleCIL & 69.6 ± 1.2 & 53.6 ± 1.4 & 64.1 ± 1.2 & 49.9 ± 1.0 & \textbf{73.2 ± 1.6} & 63.1 ± 0.8 & 66.3 ± 1.2 & 53.0 ± 1.3 & 65.6 ± 1.4 & 53.6 ± 1.6 \\
\midrule
GCN$_{\text{LLMEmb}}$ & 68.2 ± 1.3 & 40.1 ± 1.6 & 54.3 ± 1.7 & 28.7 ± 0.6 & 54.7 ± 0.8 & 31.0 ± 1.7 & 66.0 ± 1.5 & 34.2 ± 1.0 & 30.6 ± 0.9 & 0.2 ± 0.3 \\
ENGINE & 52.2 ± 1.1 & 28.3 ± 1.2 & 47.6 ± 1.1 & 25.5 ± 1.7 & 46.8 ± 1.5 & 23.7 ± 0.6 & 48.0 ± 1.2 & 21.5 ± 0.7 & 20.9 ± 1.3 & 0.1 ± 0.2 \\
GraphPrompter & 63.2 ± 1.6 & 37.3 ± 1.1 & 65.3 ± 1.5 & 34.5 ± 1.4 & 69.5 ± 0.8 & 51.6 ± 1.3 & 69.7 ± 1.4 & 51.0 ± 1.6 & 37.9 ± 1.2 & 2.2 ± 0.4 \\
GraphGPT & 62.4 ± 1.7 & 39.6 ± 0.6 & 65.0 ± 1.2 & 41.7 ± 1.3 & 71.2 ± 1.1 & 61.6 ± 1.5 & 62.2 ± 1.3 & 38.9 ± 0.7 & 43.2 ± 1.8 & 16.2 ± 1.6 \\
LLaGA  & 62.0 ± 1.2 & 39.6 ± 1.1 & 52.2 ± 0.9 & 28.0 ± 1.2 & 49.4 ± 1.3 & 30.4 ± 1.6 & 48.8 ± 1.0 & 18.0 ± 0.6 & 28.0 ± 0.7 & 0.2 ± 0.1 \\
SimGCL & \underline{78.0 ± 0.5} & \underline{67.6 ± 1.0} & \underline{78.0 ± 1.3} & \underline{63.8 ± 0.8} & 68.8 ± 1.9 & \underline{64.1 ± 1.1} & \underline{81.2 ± 1.2} & \underline{71.3 ± 0.9} & \underline{69.7 ± 1.6} & \underline{62.7 ± 1.1} \\
\midrule
\textbf{UNIT (ours) } & \textbf{81.4 ± 0.4} & \textbf{71.0 ± 0.7} & \textbf{79.7 ± 1.1} & \textbf{65.3 ± 0.6} & \underline{72.5 ± 1.2} & \textbf{65.9 ± 0.8} & \textbf{82.9 ± 1.0} & \textbf{74.6 ± 0.6} & \textbf{71.1 ± 1.2} & \textbf{64.2 ± 0.7} \\
\bottomrule
\end{tabular}
\label{few_shot}
\end{table*}

\subsection{Main Results}
The evaluation results of UNIT and the state-of-the-art baselines on five public datasets are presented in Table \ref{main_results} and the following conclusions can be derived.
First, integrating semantic information is effective to improve graph continual learning performance.
For evaluation models, the methods incorporating semantic information achieve an overall superiority, compared to traditional graph continual learning methods that merely rely on structural information.
Compared with the best traditional graph continual learning method TPP, UNIT ameliorates the classification performance up to 58.2\% in ACC$_{\text{avg}}$ and 62.3\% in ACC$_{\text{N}}$ metric.
Second, specific instruction tuning for semantic information is crucial.
LLM-based methods demonstrate competitive performance in graph continual learning tasks even without explicitly leveraging graph structure, while GLM-based methods typically design specific instructions tailored to structural information for LLM tuning.
Compared with the state-of-the-art LLM-based method for graph continual learning, UNIT achieves an average improvement of 12.8\% in ACC$_{\text{avg}}$ and 14.6\% in ACC$_{\text{N}}$ metric, which reveals the importance of specially customized tuning in the confluence of semantic-structural modeling.
Third, the proposed method UNIT unleashes LLMs potential for graph continual learning.
In general, the proposed UNIT constantly outperforms the comparison baselines across the five datasets.
The superiority demonstrates that UNIT unleashes the LLMs potential for graph continual learning.

\begin{table}[]
\setlength{\tabcolsep}{4.4pt}
\caption{Performance comparison of UNIT with variant LLM backbones on five datasets.
Note that in comparative experiments with other baseline methods, we adopt Llama3-8B as the LLM backbone of UNIT.
}
\begin{tabular}{lccccc}
\toprule
\multirow{2}{*}{Backbone} & \multicolumn{5}{c}{ACC$_\text{avg}$}                                            \\ \cmidrule{2-6} 
                         & Cora                & Citeseer                & WikiCS & Photo     &   Products              \\ \midrule
Llama3-3B       & 87.1±0.6          & 77.9±0.5           & 74.8±1.2    &  82.3±1.4        & 72.1±0.7\\
Llama3-8B       & 88.7±0.4          & 79.4±0.7           & 76.1±0.3    &  83.9±1.1        & 73.2±0.4      \\
GPT-3           & 89.5±0.3          & 80.6±1.1           & 77.2±0.6    &  84.5±0.8        & 74.3±0.5 \\
GPT-3.5       & 90.2±1.2          & 81.3±0.4           & 79.2±0.8    &  85.2±0.9        & 75.1±1.2      \\ 
\end{tabular}

\begin{tabular}{lccccc}
\toprule
\multirow{2}{*}{Backbone} & \multicolumn{5}{c}{ACC$_\text{N}$}                                           \\ \cmidrule{2-6} 
                         & Cora                & Citeseer                & WikiCS & Photo     &   Products              \\ \midrule
llama3-3B       & 84.1±0.5          & 66.9±1.1           & 64.8±1.2    &  75.3±0.6        & 61.1±0.8\\
llama3-8B       & 85.9±1.0          & 68.5±0.9          & 65.0±0.7     &  76.5±0.7        & 62.5±1.3      \\
GPT-3           & 86.7±0.7          & 69.4±1.3           & 66.1±0.4    &  77.9±1.1        & 63.2±0.8 \\
GPT-3.5       & 87.6±0.9          & 70.2±0.4           & 67.4±1.2    &  78.7±0.8        & 64.3±0.6      \\ \bottomrule
\end{tabular}
\label{backbone}
\end{table}

\subsection{Empirical Results}
\noindent{\bfseries Qualitative Results.}
To demonstrate the superiority of the proposed UNIT in each task, we conduct the additional experiment to present the accuracy performance of each task across the five datasets.
We choose the state-of-the-art methods from three baseline categories for comparative experiments, including Cosine, SimpleCIL, and SimGCL.
The results are shown in Figure \ref{task}.
Based on the observations in the figure, we make several conclusions. 
First, the proposed UNIT outperforms other methods across all tasks in the five datasets, indicating the effectiveness of unleashing LLMs potential for graph continual learning.
Second, the performance of all methods declines during backward execution, with Cosine exhibiting the most severe degradation.
This indicates that GNN-based methods struggle to achieve knowledge transfer between continual tasks with structural information only.
Third, UNIT demonstrates stable performance in long-range tasks on the Products dataset, presenting that the proposed method can effectively accomplish graph continual learning tasks. The proposed UNIT outperforms other methods across all tasks in the five datasets, indicating the effectiveness of unleashing LLMs potential for graph continual learning.

\noindent{\bfseries Few-Shot Study.}
We conduct specialized experiments to demonstrate the superiority of the proposed method UNIT in few-shot graph continual learning tasks.
The scenario configuration for few-shot graph continual learning aligns with the LLM4GCL benchmark \cite{SimGCL}, with results shown in Table \ref{few_shot}.
The results demonstrate that UNIT still outperforms most existing methods in few-shot scenarios.
Compared to the state-of-the-art methods, UNIT achieves performance improvements of up to 9.1\% in ACC$_\text{avg}$ and 14.6\% in ACC$_\text{N}$ metric, respectively.
Moreover, in few-shot scenarios, LLM-driven methods still outperform those relying solely on GNNs, even though most GNN-based methods demonstrate promising performance.
This demonstrates the effectiveness and necessity of the proposed method UNIT in unleashing the LLMs potential for graph continual learning.

\begin{figure}[th]
	\centering
    \includegraphics[width=\linewidth]{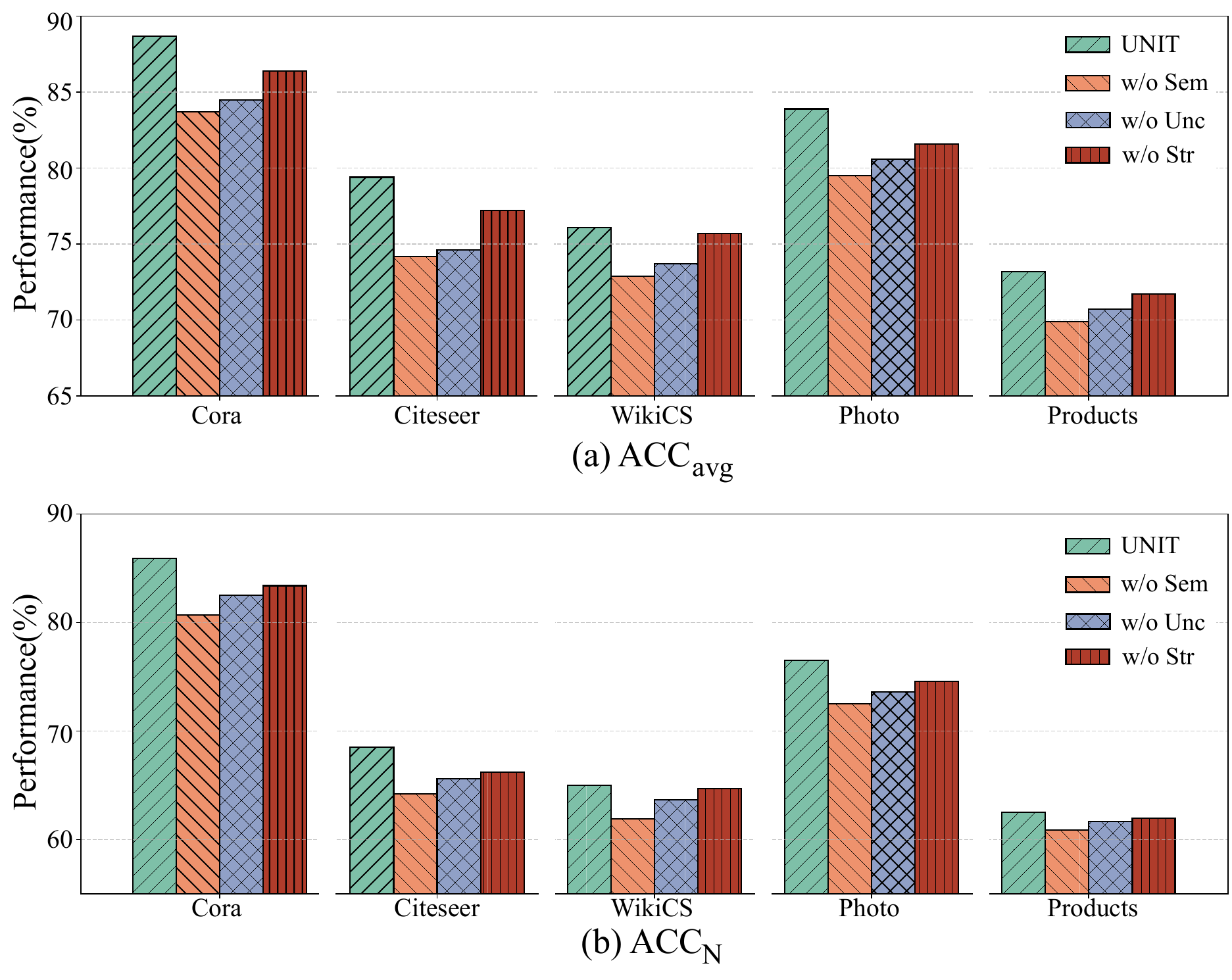}
	\caption{Ablation study results on various UNIT variants across five datasets with ACC$_{\text{avg}}$ and ACC$_{\text{N}}$ metrics. 
    }
	\label{ablation}
\end{figure}

\noindent{\bfseries Varying LLM Backbones.}
We conduct additional experiments to assess the impact of different LLM backbones on the UNIT and validate whether the capability of LLMs affect semantic comprehension.
The experiments investigate UNIT with four LLM backbones, including \emph{Llama3-3B}, \emph{Llama3-8B}, \emph{GPT-3}, and \emph{GPT-3.5}, as shown in Table \ref{backbone}.
In general, all LLM backbones improve the performance of graph continual learning task, but the extent of improvement varies between models.
Besides, it can be seen that \emph{GPT-3.5} achieves the highest performance gains on all datasets, highlighting the critical importance of LLM scale and reasoning capability in UNIT.
The proposed UNIT method can more sufficiently unleash LLMs potential for graph continual learning when it employs LLMs with more advanced semantic understanding and knowledge coverage.

\subsection{Ablation Study}
To explore how each component contributes to the performance of UNIT, we conduct a series of ablation experiments on the five datasets by excluding each module, as shown in Figure \ref{ablation}. 
This analysis aims to assess the contribution of individual modules by iteratively removing them and observing their impact on model performance.
We compare UNIT with its variants for graph continual learning:
\textbf{w/o Sem} indicates that the semantic instruction tuning module is removed, and use pre-trained LLM models without semantic fine-tuning for model training.
\textbf{w/o Unc} indicates that after semantic instruction tuning, the uncertain-aware anchor generation module is removed, and only the structural confluence modeling module is used.
\textbf{w/o Str} denotes the removal of the structural confluence modeling module after semantic instruction tuning, and only the uncertain-aware anchor generation module is used.
\textbf{UNIT} represents the use of the semantic instruction tuning module, the uncertain-aware anchor generation module, and the structural confluence modeling module, corresponding to our complete model.

\begin{figure}[th]
	\centering
    \includegraphics[width=\linewidth]{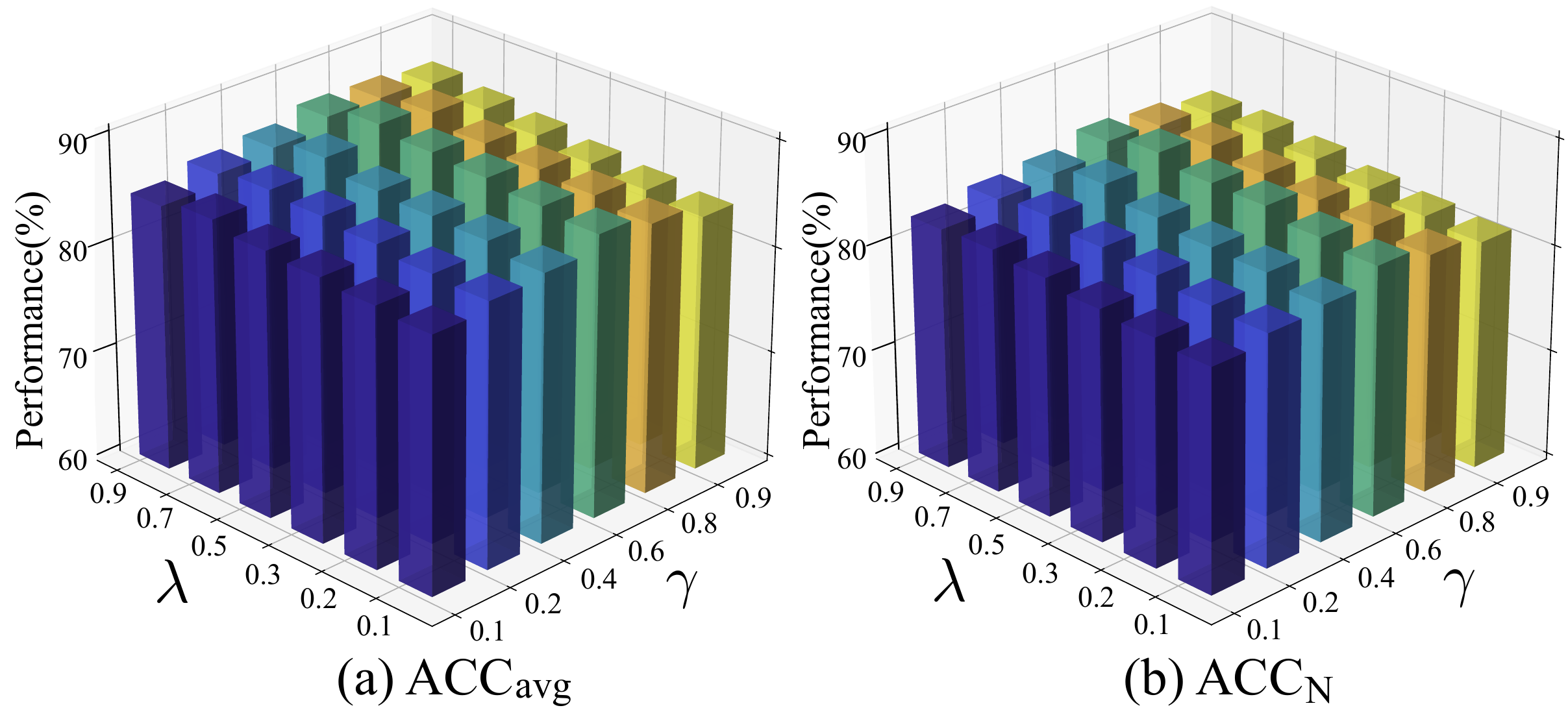}
	\caption{Parameter analysis of $\lambda$ and $\gamma$ on Cora dataset with ACC$_{\text{avg}}$ and ACC$_{\text{N}}$ metrics. Note that the values of $\lambda$ and $\gamma$ range from [0.1, $\ldots$, 0.9].
    }
	\label{sensitivity}
\end{figure}

From the results, we make several conclusions.
Firstly, we can observe that the complete model consistently outperforms its single-module counterparts in all evaluation metrics.
Besides, without specific fine-tuning of pre-trained LLM models, the classification capabilities of UNIT are compromised, indicating that specific instruction tuning for semantic information is crucial.
Moreover, it is noticeable that the uncertain-aware anchor generation module and structural confluence modeling module play critical roles in UNIT and their combination greatly benefits the graph continual learning. 
In general, each component within UNIT contributes to unleashing LLMs potential for graph continual learning.

\subsection{Parameter Analysis}
The proposed UNIT framework contains two main hyperparameters, i.e., $\lambda$ and $\gamma$, which indicate the decay factor of uncertain-aware anchor generation and structural confluence modeling, respectively.
To evaluate the robustness of UNIT under varying hyperparameter configurations, we conduct a parameter sensitivity analysis on the Cora dataset with ACC$_{\text{avg}}$ and ACC$_{\text{N}}$ metrics. 
This experiment aims to analyze the impact of the two hyper-parameters on graph continual learning performance.
Specifically, we vary the values of $\lambda$ and $\gamma$ within the range of
[0.1, $\ldots$, 0.9] and Figure \ref{sensitivity} presents the experimental results.
Based on the observations in the figure, we make several conclusions.
Firstly, UNIT tends to exhibit suboptimal performance when $\lambda$ and $\gamma$ are set to low values, e.g., 0.1 and 0.2.
This emphasizes the critical role historical knowledge plays within the UNIT framework and highlights its effectiveness.
Secondly, we can observe that excessively high values of $\lambda$ and $\gamma$ also adversely affect performance, as they may obscure the importance of new knowledge.
Finally, UNIT demonstrates relatively stable ACC$_{\text{avg}}$ and ACC$_{\text{N}}$ performance across a wide range of $\lambda$ and $\gamma$ values, which proves its robustness.

\section{Conclusion}
In this paper, we investigate the systematic exploration of LLMs based on graph continual learning tasks and propose a novel framework, Unleash Large Language Models Potential for Graph Continual Learning (UNIT).
Accordingly, we design specific prompt instructions to fine-tune pre-trained LLMs to facilitate the LLM comprehension ability to graph continual learning tasks.
Moreover, UNIT proposes uncertainty-aware anchor generation and structural confluence modeling to achieve comprehensive knowledge transfer and sufficient integration of semantic and structural information.
The experimental results show that UNIT can effectively accomplish the graph continual learning task, which verifies the its effectiveness and robustness.

\begin{acks}
The work is supported by the National Natural Science Foundation of China (No. 62272487 and 62433016).
\end{acks}
\bibliographystyle{ACM-Reference-Format}
\bibliography{acmart}


\begin{thebibliography}{43}


\ifx \showCODEN    \undefined \def \showCODEN     #1{\unskip}     \fi
\ifx \showISBNx    \undefined \def \showISBNx     #1{\unskip}     \fi
\ifx \showISBNxiii \undefined \def \showISBNxiii  #1{\unskip}     \fi
\ifx \showISSN     \undefined \def \showISSN      #1{\unskip}     \fi
\ifx \showLCCN     \undefined \def \showLCCN      #1{\unskip}     \fi
\ifx \shownote     \undefined \def \shownote      #1{#1}          \fi
\ifx \showarticletitle \undefined \def \showarticletitle #1{#1}   \fi
\ifx \showURL      \undefined \def \showURL       {\relax}        \fi
\providecommand\bibfield[2]{#2}
\providecommand\bibinfo[2]{#2}
\providecommand\natexlab[1]{#1}
\providecommand\showeprint[2][]{arXiv:#2}

\bibitem[Brown et~al\mbox{.}(2020)]%
        {few_1}
\bibfield{author}{\bibinfo{person}{Tom Brown}, \bibinfo{person}{Benjamin Mann}, \bibinfo{person}{Nick Ryder}, \bibinfo{person}{Melanie Subbiah}, \bibinfo{person}{Jared~D Kaplan}, \bibinfo{person}{Prafulla Dhariwal}, \bibinfo{person}{Arvind Neelakantan}, \bibinfo{person}{Pranav Shyam}, \bibinfo{person}{Girish Sastry}, \bibinfo{person}{Amanda Askell}, {et~al\mbox{.}}} \bibinfo{year}{2020}\natexlab{}.
\newblock \showarticletitle{Language models are few-shot learners}.
\newblock \bibinfo{journal}{\emph{Advances in Neural Information Processing Systems}}  \bibinfo{volume}{33} (\bibinfo{year}{2020}), \bibinfo{pages}{1877--1901}.
\newblock


\bibitem[Chen et~al\mbox{.}(2025)]%
        {social}
\bibfield{author}{\bibinfo{person}{Qiang Chen}, \bibinfo{person}{Zhongze Wu}, \bibinfo{person}{Xiu Su}, \bibinfo{person}{Xi Lin}, \bibinfo{person}{Zhe Qu}, \bibinfo{person}{Shan You}, \bibinfo{person}{Shuo Yang}, {and} \bibinfo{person}{Chang Xu}.} \bibinfo{year}{2025}\natexlab{}.
\newblock \showarticletitle{Stable Fair Graph Representation Learning with Lipschitz Constraint}. In \bibinfo{booktitle}{\emph{Proceedings of the International Conference on Machine Learning}}.
\newblock


\bibitem[Chen et~al\mbox{.}(2024b)]%
        {Llaga}
\bibfield{author}{\bibinfo{person}{Runjin Chen}, \bibinfo{person}{Tong Zhao}, \bibinfo{person}{Ajay Jaiswal}, \bibinfo{person}{Neil Shah}, {and} \bibinfo{person}{Zhangyang Wang}.} \bibinfo{year}{2024}\natexlab{b}.
\newblock \showarticletitle{Llaga: Large language and graph assistant}.
\newblock \bibinfo{journal}{\emph{arXiv preprint arXiv:2402.08170}} (\bibinfo{year}{2024}).
\newblock


\bibitem[Chen et~al\mbox{.}(2024a)]%
        {Citeseer}
\bibfield{author}{\bibinfo{person}{Zhikai Chen}, \bibinfo{person}{Haitao Mao}, \bibinfo{person}{Hang Li}, \bibinfo{person}{Wei Jin}, \bibinfo{person}{Hongzhi Wen}, \bibinfo{person}{Xiaochi Wei}, \bibinfo{person}{Shuaiqiang Wang}, \bibinfo{person}{Dawei Yin}, \bibinfo{person}{Wenqi Fan}, \bibinfo{person}{Hui Liu}, {et~al\mbox{.}}} \bibinfo{year}{2024}\natexlab{a}.
\newblock \showarticletitle{Exploring the potential of large language models (llms) in learning on graphs}.
\newblock \bibinfo{journal}{\emph{ACM SIGKDD Explorations Newsletter}} \bibinfo{volume}{25}, \bibinfo{number}{2} (\bibinfo{year}{2024}), \bibinfo{pages}{42--61}.
\newblock


\bibitem[Cheng et~al\mbox{.}(2025)]%
        {SimGCL}
\bibfield{author}{\bibinfo{person}{Ziyang Cheng}, \bibinfo{person}{Zhixun Li}, \bibinfo{person}{Yuhan Li}, \bibinfo{person}{Yixin Song}, \bibinfo{person}{Kangyi Zhao}, \bibinfo{person}{Dawei Cheng}, \bibinfo{person}{Jia Li}, {and} \bibinfo{person}{Jeffrey~Xu Yu}.} \bibinfo{year}{2025}\natexlab{}.
\newblock \showarticletitle{Can LLMs Alleviate Catastrophic Forgetting in Graph Continual Learning? A Systematic Study}.
\newblock \bibinfo{journal}{\emph{arXiv preprint arXiv:2505.18697}} (\bibinfo{year}{2025}).
\newblock


\bibitem[Choi et~al\mbox{.}(2024)]%
        {GCL}
\bibfield{author}{\bibinfo{person}{Seungyoon Choi}, \bibinfo{person}{Wonjoong Kim}, \bibinfo{person}{Sungwon Kim}, \bibinfo{person}{Yeonjun In}, \bibinfo{person}{Sein Kim}, {and} \bibinfo{person}{Chanyoung Park}.} \bibinfo{year}{2024}\natexlab{}.
\newblock \showarticletitle{DSLR: diversity enhancement and structure learning for rehearsal-based graph continual learning}. In \bibinfo{booktitle}{\emph{Proceedings of the ACM Web Conference 2024}}. \bibinfo{pages}{733--744}.
\newblock


\bibitem[Devlin et~al\mbox{.}(2019)]%
        {BERT}
\bibfield{author}{\bibinfo{person}{Jacob Devlin}, \bibinfo{person}{Ming-Wei Chang}, \bibinfo{person}{Kenton Lee}, {and} \bibinfo{person}{Kristina Toutanova}.} \bibinfo{year}{2019}\natexlab{}.
\newblock \showarticletitle{Bert: Pre-training of deep bidirectional transformers for language understanding}. In \bibinfo{booktitle}{\emph{Proceedings of the Conference of the North American Chapter of the Association for Computational Linguistics: human language technologies, volume 1 (long and short papers)}}. \bibinfo{pages}{4171--4186}.
\newblock


\bibitem[Han et~al\mbox{.}(2025)]%
        {Financial}
\bibfield{author}{\bibinfo{person}{Xiaolin Han}, \bibinfo{person}{Yikun Zhang}, \bibinfo{person}{Chenhao Ma}, \bibinfo{person}{Lingyun Song}, \bibinfo{person}{Reynold Cheng}, {and} \bibinfo{person}{Xuequn Shang}.} \bibinfo{year}{2025}\natexlab{}.
\newblock \showarticletitle{TempASD: Temporal Anomalous Subgraph Discovery in Large-Scale Dynamic Financial Networks}. In \bibinfo{booktitle}{\emph{Proceedings of the ACM SIGKDD Conference on Knowledge Discovery and Data Mining V. 2}}. \bibinfo{pages}{826--837}.
\newblock


\bibitem[He et~al\mbox{.}(2023)]%
        {Cora}
\bibfield{author}{\bibinfo{person}{Xiaoxin He}, \bibinfo{person}{Xavier Bresson}, \bibinfo{person}{Thomas Laurent}, \bibinfo{person}{Adam Perold}, \bibinfo{person}{Yann LeCun}, {and} \bibinfo{person}{Bryan Hooi}.} \bibinfo{year}{2023}\natexlab{}.
\newblock \showarticletitle{Harnessing explanations: Llm-to-lm interpreter for enhanced text-attributed graph representation learning}.
\newblock \bibinfo{journal}{\emph{arXiv preprint arXiv:2305.19523}} (\bibinfo{year}{2023}).
\newblock


\bibitem[Hu et~al\mbox{.}(2022)]%
        {lora}
\bibfield{author}{\bibinfo{person}{Edward~J Hu}, \bibinfo{person}{Yelong Shen}, \bibinfo{person}{Phillip Wallis}, \bibinfo{person}{Zeyuan Allen-Zhu}, \bibinfo{person}{Yuanzhi Li}, \bibinfo{person}{Shean Wang}, \bibinfo{person}{Lu Wang}, \bibinfo{person}{Weizhu Chen}, {et~al\mbox{.}}} \bibinfo{year}{2022}\natexlab{}.
\newblock \showarticletitle{Lora: Low-rank adaptation of large language models.}. In \bibinfo{booktitle}{\emph{International Conference on Learning Representations}}. \bibinfo{pages}{3}.
\newblock


\bibitem[Hu et~al\mbox{.}(2020)]%
        {Products}
\bibfield{author}{\bibinfo{person}{Weihua Hu}, \bibinfo{person}{Matthias Fey}, \bibinfo{person}{Marinka Zitnik}, \bibinfo{person}{Yuxiao Dong}, \bibinfo{person}{Hongyu Ren}, \bibinfo{person}{Bowen Liu}, \bibinfo{person}{Michele Catasta}, {and} \bibinfo{person}{Jure Leskovec}.} \bibinfo{year}{2020}\natexlab{}.
\newblock \showarticletitle{Open graph benchmark: Datasets for machine learning on graphs}.
\newblock \bibinfo{journal}{\emph{Advances in Neural Information Processing Systems}}  \bibinfo{volume}{33} (\bibinfo{year}{2020}), \bibinfo{pages}{22118--22133}.
\newblock


\bibitem[Huang et~al\mbox{.}(2025)]%
        {social_2}
\bibfield{author}{\bibinfo{person}{Huang}, \bibinfo{person}{Wang Tairan}, \bibinfo{person}{Li Yili}, \bibinfo{person}{He Qiutong}, \bibinfo{person}{Gao Changlong}, {and} \bibinfo{person}{Jianliang}.} \bibinfo{year}{2025}\natexlab{}.
\newblock \showarticletitle{Can LLMs Find Fraudsters? Multi-level LLM Enhanced Graph Fraud Detection}. In \bibinfo{booktitle}{\emph{Proceedings of the ACM International Conference on Multimedia}}.
\newblock


\bibitem[Hung et~al\mbox{.}(2019)]%
        {Re}
\bibfield{author}{\bibinfo{person}{Ching-Yi Hung}, \bibinfo{person}{Cheng-Hao Tu}, \bibinfo{person}{Cheng-En Wu}, \bibinfo{person}{Chien-Hung Chen}, \bibinfo{person}{Yi-Ming Chan}, {and} \bibinfo{person}{Chu-Song Chen}.} \bibinfo{year}{2019}\natexlab{}.
\newblock \showarticletitle{Compacting, picking and growing for unforgetting continual learning}.
\newblock \bibinfo{journal}{\emph{Advances in Neural Information Processing Systems}}  \bibinfo{volume}{32} (\bibinfo{year}{2019}).
\newblock


\bibitem[Jia et~al\mbox{.}(2025)]%
        {Recommend}
\bibfield{author}{\bibinfo{person}{Pengyue Jia}, \bibinfo{person}{Jingtong Gao}, \bibinfo{person}{Yejing Wang}, \bibinfo{person}{Yuhao Wang}, \bibinfo{person}{Xiaopeng Li}, \bibinfo{person}{Qidong Liu}, \bibinfo{person}{Yichao Wang}, \bibinfo{person}{Bo Chen}, \bibinfo{person}{Huifeng Guo}, {and} \bibinfo{person}{Ruiming Tang}.} \bibinfo{year}{2025}\natexlab{}.
\newblock \showarticletitle{Joint Modeling in Deep Recommender Systems}. In \bibinfo{booktitle}{\emph{Companion Proceedings of the ACM on Web Conference}}. \bibinfo{pages}{17--20}.
\newblock


\bibitem[Kipf and Welling(2017)]%
        {GCN}
\bibfield{author}{\bibinfo{person}{Thomas~N Kipf} {and} \bibinfo{person}{Max Welling}.} \bibinfo{year}{2017}\natexlab{}.
\newblock \showarticletitle{Semi-supervised classification with graph convolutional networks}. In \bibinfo{booktitle}{\emph{International Conference on Learning Representations}}. \bibinfo{pages}{1031--1049}.
\newblock


\bibitem[Kirkpatrick et~al\mbox{.}(2017)]%
        {EWC}
\bibfield{author}{\bibinfo{person}{James Kirkpatrick}, \bibinfo{person}{Razvan Pascanu}, \bibinfo{person}{Neil Rabinowitz}, \bibinfo{person}{Joel Veness}, \bibinfo{person}{Guillaume Desjardins}, \bibinfo{person}{Andrei~A Rusu}, \bibinfo{person}{Kieran Milan}, \bibinfo{person}{John Quan}, \bibinfo{person}{Tiago Ramalho}, \bibinfo{person}{Agnieszka Grabska-Barwinska}, {et~al\mbox{.}}} \bibinfo{year}{2017}\natexlab{}.
\newblock \showarticletitle{Overcoming catastrophic forgetting in neural networks}.
\newblock \bibinfo{journal}{\emph{Proceedings of the National Academy of Sciences}} \bibinfo{volume}{114}, \bibinfo{number}{13} (\bibinfo{year}{2017}), \bibinfo{pages}{3521--3526}.
\newblock


\bibitem[Li et~al\mbox{.}(2024)]%
        {Glbench}
\bibfield{author}{\bibinfo{person}{Yuhan Li}, \bibinfo{person}{Peisong Wang}, \bibinfo{person}{Xiao Zhu}, \bibinfo{person}{Aochuan Chen}, \bibinfo{person}{Haiyun Jiang}, \bibinfo{person}{Deng Cai}, \bibinfo{person}{Victor~W Chan}, {and} \bibinfo{person}{Jia Li}.} \bibinfo{year}{2024}\natexlab{}.
\newblock \showarticletitle{Glbench: A comprehensive benchmark for graph with large language models}.
\newblock \bibinfo{journal}{\emph{Advances in Neural Information Processing Systems}}  \bibinfo{volume}{37} (\bibinfo{year}{2024}), \bibinfo{pages}{42349--42368}.
\newblock


\bibitem[Li and Hoiem(2017)]%
        {LwF}
\bibfield{author}{\bibinfo{person}{Zhizhong Li} {and} \bibinfo{person}{Derek Hoiem}.} \bibinfo{year}{2017}\natexlab{}.
\newblock \showarticletitle{Learning without forgetting}.
\newblock \bibinfo{journal}{\emph{IEEE Transactions on Pattern Analysis and Machine Intelligence}} \bibinfo{volume}{40}, \bibinfo{number}{12} (\bibinfo{year}{2017}), \bibinfo{pages}{2935--2947}.
\newblock


\bibitem[Liang and Li(2024)]%
        {Inflora}
\bibfield{author}{\bibinfo{person}{Yan-Shuo Liang} {and} \bibinfo{person}{Wu-Jun Li}.} \bibinfo{year}{2024}\natexlab{}.
\newblock \showarticletitle{Inflora: Interference-free low-rank adaptation for continual learning}. In \bibinfo{booktitle}{\emph{Proceedings of the IEEE/CVF Conference on Computer Vision and Pattern Recognition}}. \bibinfo{pages}{23638--23647}.
\newblock


\bibitem[Liu et~al\mbox{.}(2023)]%
        {WikiCS}
\bibfield{author}{\bibinfo{person}{Hao Liu}, \bibinfo{person}{Jiarui Feng}, \bibinfo{person}{Lecheng Kong}, \bibinfo{person}{Ningyue Liang}, \bibinfo{person}{Dacheng Tao}, \bibinfo{person}{Yixin Chen}, {and} \bibinfo{person}{Muhan Zhang}.} \bibinfo{year}{2023}\natexlab{}.
\newblock \showarticletitle{One for all: Towards training one graph model for all classification tasks}.
\newblock \bibinfo{journal}{\emph{arXiv preprint arXiv:2310.00149}} (\bibinfo{year}{2023}).
\newblock


\bibitem[Liu et~al\mbox{.}(2019)]%
        {Roberta}
\bibfield{author}{\bibinfo{person}{Yinhan Liu}, \bibinfo{person}{Myle Ott}, \bibinfo{person}{Naman Goyal}, \bibinfo{person}{Jingfei Du}, \bibinfo{person}{Mandar Joshi}, \bibinfo{person}{Danqi Chen}, \bibinfo{person}{Omer Levy}, \bibinfo{person}{Mike Lewis}, \bibinfo{person}{Luke Zettlemoyer}, {and} \bibinfo{person}{Veselin Stoyanov}.} \bibinfo{year}{2019}\natexlab{}.
\newblock \showarticletitle{Roberta: A robustly optimized bert pretraining approach}.
\newblock \bibinfo{journal}{\emph{arXiv preprint arXiv:1907.11692}} (\bibinfo{year}{2019}).
\newblock


\bibitem[Liu et~al\mbox{.}(2024)]%
        {GraphPrompter}
\bibfield{author}{\bibinfo{person}{Zheyuan Liu}, \bibinfo{person}{Xiaoxin He}, \bibinfo{person}{Yijun Tian}, {and} \bibinfo{person}{Nitesh~V Chawla}.} \bibinfo{year}{2024}\natexlab{}.
\newblock \showarticletitle{Can we soft prompt llms for graph learning tasks?}. In \bibinfo{booktitle}{\emph{Companion Proceedings of the ACM Web Conference}}. \bibinfo{pages}{481--484}.
\newblock


\bibitem[Ma and Tang(2021)]%
        {TAG}
\bibfield{author}{\bibinfo{person}{Yao Ma} {and} \bibinfo{person}{Jiliang Tang}.} \bibinfo{year}{2021}\natexlab{}.
\newblock \bibinfo{booktitle}{\emph{Deep learning on graphs}}.
\newblock \bibinfo{publisher}{Cambridge University Press}.
\newblock


\bibitem[Niu et~al\mbox{.}(2024)]%
        {Cosine}
\bibfield{author}{\bibinfo{person}{Chaoxi Niu}, \bibinfo{person}{Guansong Pang}, \bibinfo{person}{Ling Chen}, {and} \bibinfo{person}{Bing Liu}.} \bibinfo{year}{2024}\natexlab{}.
\newblock \showarticletitle{Replay-and-forget-free graph class-incremental learning: A task profiling and prompting approach}.
\newblock \bibinfo{journal}{\emph{Advances in Neural Information Processing Systems}}  \bibinfo{volume}{37} (\bibinfo{year}{2024}), \bibinfo{pages}{87978--88002}.
\newblock


\bibitem[Rebuffi et~al\mbox{.}(2017)]%
        {GCL_bench}
\bibfield{author}{\bibinfo{person}{Sylvestre-Alvise Rebuffi}, \bibinfo{person}{Hakan Bilen}, {and} \bibinfo{person}{Andrea Vedaldi}.} \bibinfo{year}{2017}\natexlab{}.
\newblock \showarticletitle{Learning multiple visual domains with residual adapters}.
\newblock \bibinfo{journal}{\emph{Advances in Neural Information Processing Systems}}  \bibinfo{volume}{30} (\bibinfo{year}{2017}).
\newblock


\bibitem[Tan et~al\mbox{.}(2025)]%
        {CoT}
\bibfield{author}{\bibinfo{person}{Yanchao Tan}, \bibinfo{person}{Hang Lv}, \bibinfo{person}{Pengxiang Zhan}, \bibinfo{person}{Shiping Wang}, {and} \bibinfo{person}{Carl Yang}.} \bibinfo{year}{2025}\natexlab{}.
\newblock \showarticletitle{Graph-oriented Instruction Tuning of Large Language Models for Generic Graph Mining}.
\newblock \bibinfo{journal}{\emph{IEEE Transactions on Pattern Analysis and Machine Intelligence}} (\bibinfo{year}{2025}).
\newblock


\bibitem[Tang et~al\mbox{.}(2024)]%
        {Graphgpt}
\bibfield{author}{\bibinfo{person}{Jiabin Tang}, \bibinfo{person}{Yuhao Yang}, \bibinfo{person}{Wei Wei}, \bibinfo{person}{Lei Shi}, \bibinfo{person}{Lixin Su}, \bibinfo{person}{Suqi Cheng}, \bibinfo{person}{Dawei Yin}, {and} \bibinfo{person}{Chao Huang}.} \bibinfo{year}{2024}\natexlab{}.
\newblock \showarticletitle{Graphgpt: Graph instruction tuning for large language models}. In \bibinfo{booktitle}{\emph{Proceedings of the 47th International ACM SIGIR Conference on Research and Development in Information Retrieval}}. \bibinfo{pages}{491--500}.
\newblock


\bibitem[Touvron et~al\mbox{.}(2023)]%
        {llama}
\bibfield{author}{\bibinfo{person}{Hugo Touvron}, \bibinfo{person}{Louis Martin}, \bibinfo{person}{Kevin Stone}, \bibinfo{person}{Peter Albert}, \bibinfo{person}{Amjad Almahairi}, \bibinfo{person}{Yasmine Babaei}, \bibinfo{person}{Nikolay Bashlykov}, \bibinfo{person}{Soumya Batra}, \bibinfo{person}{Prajjwal Bhargava}, \bibinfo{person}{Shruti Bhosale}, {et~al\mbox{.}}} \bibinfo{year}{2023}\natexlab{}.
\newblock \showarticletitle{Llama 2: Open foundation and fine-tuned chat models}.
\newblock \bibinfo{journal}{\emph{arXiv preprint arXiv:2307.09288}} (\bibinfo{year}{2023}).
\newblock


\bibitem[Wang et~al\mbox{.}(2024)]%
        {TEA-GLM}
\bibfield{author}{\bibinfo{person}{Duo Wang}, \bibinfo{person}{Yuan Zuo}, \bibinfo{person}{Fengzhi Li}, {and} \bibinfo{person}{Junjie Wu}.} \bibinfo{year}{2024}\natexlab{}.
\newblock \showarticletitle{Llms as zero-shot graph learners: Alignment of gnn representations with llm token embeddings}.
\newblock \bibinfo{journal}{\emph{Advances in Neural Information Processing Systems}}  \bibinfo{volume}{37} (\bibinfo{year}{2024}), \bibinfo{pages}{5950--5973}.
\newblock


\bibitem[Wang et~al\mbox{.}(2023)]%
        {TPP}
\bibfield{author}{\bibinfo{person}{Qi-Wei Wang}, \bibinfo{person}{Da-Wei Zhou}, \bibinfo{person}{Yi-Kai Zhang}, \bibinfo{person}{De-Chuan Zhan}, {and} \bibinfo{person}{Han-Jia Ye}.} \bibinfo{year}{2023}\natexlab{}.
\newblock \showarticletitle{Few-shot class-incremental learning via training-free prototype calibration}.
\newblock \bibinfo{journal}{\emph{Advances in Neural Information Processing Systems}}  \bibinfo{volume}{36} (\bibinfo{year}{2023}), \bibinfo{pages}{15060--15076}.
\newblock


\bibitem[Wu et~al\mbox{.}(2025)]%
        {GCN_emb}
\bibfield{author}{\bibinfo{person}{Xixi Wu}, \bibinfo{person}{Yifei Shen}, \bibinfo{person}{Fangzhou Ge}, \bibinfo{person}{Caihua Shan}, \bibinfo{person}{Yizhu Jiao}, \bibinfo{person}{Xiangguo Sun}, {and} \bibinfo{person}{Hong Cheng}.} \bibinfo{year}{2025}\natexlab{}.
\newblock \showarticletitle{A comprehensive analysis on llm-based node classification algorithms}.
\newblock \bibinfo{journal}{\emph{arXiv preprint arXiv:2502.00829}} (\bibinfo{year}{2025}).
\newblock


\bibitem[Yan et~al\mbox{.}(2023)]%
        {Photo}
\bibfield{author}{\bibinfo{person}{Hao Yan}, \bibinfo{person}{Chaozhuo Li}, \bibinfo{person}{Ruosong Long}, \bibinfo{person}{Chao Yan}, \bibinfo{person}{Jianan Zhao}, \bibinfo{person}{Wenwen Zhuang}, \bibinfo{person}{Jun Yin}, \bibinfo{person}{Peiyan Zhang}, \bibinfo{person}{Weihao Han}, \bibinfo{person}{Hao Sun}, {et~al\mbox{.}}} \bibinfo{year}{2023}\natexlab{}.
\newblock \showarticletitle{A comprehensive study on text-attributed graphs: Benchmarking and rethinking}.
\newblock \bibinfo{journal}{\emph{Advances in Neural Information Processing Systems}}  \bibinfo{volume}{36} (\bibinfo{year}{2023}), \bibinfo{pages}{17238--17264}.
\newblock


\bibitem[Yan et~al\mbox{.}(2025)]%
        {MAGRL}
\bibfield{author}{\bibinfo{person}{Hao Yan}, \bibinfo{person}{Chaozhuo Li}, \bibinfo{person}{Jun Yin}, \bibinfo{person}{Zhigang Yu}, \bibinfo{person}{Weihao Han}, \bibinfo{person}{Mingzheng Li}, \bibinfo{person}{Zhengxin Zeng}, \bibinfo{person}{Hao Sun}, {and} \bibinfo{person}{Senzhang Wang}.} \bibinfo{year}{2025}\natexlab{}.
\newblock \showarticletitle{When Graph Meets Multimodal: Benchmarking and Meditating on Multimodal Attributed Graph Learning}. In \bibinfo{booktitle}{\emph{Proceedings of the ACM SIGKDD Conference on Knowledge Discovery and Data Mining V. 2}}. \bibinfo{pages}{5842--5853}.
\newblock


\bibitem[Zhang et~al\mbox{.}(2024)]%
        {Graphtranslator}
\bibfield{author}{\bibinfo{person}{Mengmei Zhang}, \bibinfo{person}{Mingwei Sun}, \bibinfo{person}{Peng Wang}, \bibinfo{person}{Shen Fan}, \bibinfo{person}{Yanhu Mo}, \bibinfo{person}{Xiaoxiao Xu}, \bibinfo{person}{Hong Liu}, \bibinfo{person}{Cheng Yang}, {and} \bibinfo{person}{Chuan Shi}.} \bibinfo{year}{2024}\natexlab{}.
\newblock \showarticletitle{Graphtranslator: Aligning graph model to large language model for open-ended tasks}. In \bibinfo{booktitle}{\emph{Proceedings of the ACM Web Conference 2024}}. \bibinfo{pages}{1003--1014}.
\newblock


\bibitem[Zhang et~al\mbox{.}(2023b)]%
        {is}
\bibfield{author}{\bibinfo{person}{Peiyan Zhang}, \bibinfo{person}{Yuchen Yan}, \bibinfo{person}{Chaozhuo Li}, \bibinfo{person}{Senzhang Wang}, \bibinfo{person}{Xing Xie}, \bibinfo{person}{Guojie Song}, {and} \bibinfo{person}{Sunghun Kim}.} \bibinfo{year}{2023}\natexlab{b}.
\newblock \showarticletitle{Continual learning on dynamic graphs via parameter isolation}. In \bibinfo{booktitle}{\emph{Proceedings of the international ACM SIGIR conference on research and development in information retrieval}}. \bibinfo{pages}{601--611}.
\newblock


\bibitem[Zhang et~al\mbox{.}(2025)]%
        {Locle}
\bibfield{author}{\bibinfo{person}{Taiyan Zhang}, \bibinfo{person}{Renchi Yang}, \bibinfo{person}{Yurui Lai}, \bibinfo{person}{Mingyu Yan}, \bibinfo{person}{Xiaochun Ye}, {and} \bibinfo{person}{Dongrui Fan}.} \bibinfo{year}{2025}\natexlab{}.
\newblock \showarticletitle{Leveraging large language models for effective label-free node classification in text-attributed graphs}. In \bibinfo{booktitle}{\emph{Proceedings of the International ACM SIGIR Conference on Research and Development in Information Retrieval}}. \bibinfo{pages}{698--708}.
\newblock


\bibitem[Zhang et~al\mbox{.}(2022a)]%
        {is_2}
\bibfield{author}{\bibinfo{person}{Xikun Zhang}, \bibinfo{person}{Dongjin Song}, {and} \bibinfo{person}{Dacheng Tao}.} \bibinfo{year}{2022}\natexlab{a}.
\newblock \showarticletitle{Hierarchical prototype networks for continual graph representation learning}.
\newblock \bibinfo{journal}{\emph{IEEE Transactions on Pattern Analysis and Machine Intelligence}} \bibinfo{volume}{45}, \bibinfo{number}{4} (\bibinfo{year}{2022}), \bibinfo{pages}{4622--4636}.
\newblock


\bibitem[Zhang et~al\mbox{.}(2022b)]%
        {rh}
\bibfield{author}{\bibinfo{person}{Xikun Zhang}, \bibinfo{person}{Dongjin Song}, {and} \bibinfo{person}{Dacheng Tao}.} \bibinfo{year}{2022}\natexlab{b}.
\newblock \showarticletitle{Sparsified subgraph memory for continual graph representation learning}. In \bibinfo{booktitle}{\emph{2022 IEEE International Conference on Data Mining}}. \bibinfo{pages}{1335--1340}.
\newblock


\bibitem[Zhang et~al\mbox{.}(2023a)]%
        {rh_2}
\bibfield{author}{\bibinfo{person}{Xikun Zhang}, \bibinfo{person}{Dongjin Song}, {and} \bibinfo{person}{Dacheng Tao}.} \bibinfo{year}{2023}\natexlab{a}.
\newblock \showarticletitle{Ricci curvature-based graph sparsification for continual graph representation learning}.
\newblock \bibinfo{journal}{\emph{IEEE Transactions on Neural Networks and Learning Systems}} (\bibinfo{year}{2023}).
\newblock


\bibitem[Zhou et~al\mbox{.}(2025)]%
        {SimpleCIL}
\bibfield{author}{\bibinfo{person}{Da-Wei Zhou}, \bibinfo{person}{Zi-Wen Cai}, \bibinfo{person}{Han-Jia Ye}, \bibinfo{person}{De-Chuan Zhan}, {and} \bibinfo{person}{Ziwei Liu}.} \bibinfo{year}{2025}\natexlab{}.
\newblock \showarticletitle{Revisiting class-incremental learning with pre-trained models: Generalizability and adaptivity are all you need}.
\newblock \bibinfo{journal}{\emph{International Journal of Computer Vision}} \bibinfo{volume}{133}, \bibinfo{number}{3} (\bibinfo{year}{2025}), \bibinfo{pages}{1012--1032}.
\newblock


\bibitem[Zhou and Cao(2021)]%
        {Re_2}
\bibfield{author}{\bibinfo{person}{Fan Zhou} {and} \bibinfo{person}{Chengtai Cao}.} \bibinfo{year}{2021}\natexlab{}.
\newblock \showarticletitle{Overcoming catastrophic forgetting in graph neural networks with experience replay}. In \bibinfo{booktitle}{\emph{Proceedings of the AAAI conference on artificial intelligence}}, Vol.~\bibinfo{volume}{35}. \bibinfo{pages}{4714--4722}.
\newblock


\bibitem[Zhu et~al\mbox{.}(2025)]%
        {BiLoRA}
\bibfield{author}{\bibinfo{person}{Hao Zhu}, \bibinfo{person}{Yifei Zhang}, \bibinfo{person}{Junhao Dong}, {and} \bibinfo{person}{Piotr Koniusz}.} \bibinfo{year}{2025}\natexlab{}.
\newblock \showarticletitle{BiLoRA: Almost-Orthogonal Parameter Spaces for Continual Learning}. In \bibinfo{booktitle}{\emph{Proceedings of the Computer Vision and Pattern Recognition Conference}}. \bibinfo{pages}{25613--25622}.
\newblock


\bibitem[Zhu et~al\mbox{.}(2024)]%
        {ENGINE}
\bibfield{author}{\bibinfo{person}{Yun Zhu}, \bibinfo{person}{Yaoke Wang}, \bibinfo{person}{Haizhou Shi}, {and} \bibinfo{person}{Siliang Tang}.} \bibinfo{year}{2024}\natexlab{}.
\newblock \showarticletitle{Efficient tuning and inference for large language models on textual graphs}.
\newblock \bibinfo{journal}{\emph{arXiv preprint arXiv:2401.15569}} (\bibinfo{year}{2024}).
\newblock


\end{thebibliography}

\end{document}